\pdfoutput=1

\documentclass[11pt]{article}

\usepackage{authblk}
\usepackage{acl}

\usepackage{times}
\usepackage{latexsym}

\usepackage{enumitem}
\usepackage{amsmath}
\usepackage{graphicx}
\usepackage{array}
\usepackage{CJKutf8}
\usepackage{multirow}
\usepackage{makecell}

\usepackage[T1]{fontenc}

\usepackage[utf8]{inputenc}

\usepackage{microtype}

\usepackage[hang]{footmisc}
\title{Cross-lingual Inference with A Chinese Entailment Graph}

\author{{\bf Tianyi Li$^{\diamond}$} \quad {\bf Sabine Weber$^{\diamond}$} \quad {\bf Mohammad Javad Hosseini$^{\diamond}$\thanks{Now at Google Research.}} \\ \quad {\bf Liane Guillou $^{\diamond}$} \quad {\bf Mark Steedman$^{\diamond}$} \\
        $^\diamond$ School of Informatics, University of Edinburgh \\
        \texttt{tianyi.li@ed.ac.uk, s.weber@sms.ed.ac.uk} \\
       \texttt{javad.hosseini@ed.ac.uk},
    \texttt{\{lguillou, steedman\}@inf.ed.ac.uk}
}

\begin{document}
\maketitle
\begin{CJK*}{UTF8}{gbsn}
\begin{abstract}
Predicate entailment detection is a crucial task for question-answering from text, where previous work has explored unsupervised learning of entailment graphs from typed open relation triples. In this paper, we present the first pipeline for building Chinese entailment graphs, which involves a novel high-recall open relation extraction (ORE) method and the first Chinese fine-grained entity typing dataset under the FIGER type ontology. Through experiments on the Levy-Holt dataset, we verify the strength of our Chinese entailment graph, and reveal the cross-lingual complementarity: on the parallel Levy-Holt dataset, an ensemble of Chinese and English entailment graphs outperforms both monolingual graphs, and raises unsupervised SOTA by 4.7 AUC points.
\end{abstract}

\section{Introduction}
\label{Sec:Intro}

Predicate entailment detection is important for many tasks of natural language understanding (NLU), including reading comprehension and semantic parsing. Suppose we wish to answer a question by finding a relation $\textbf{V}$ between entities $\textbf{A}$ and $\textbf{B}$. Often, $\textbf{V}$ cannot be found directly from the reference passage or database, but another relation $\textbf{U}$ can be found between $\textbf{A}$ and $\textbf{B}$, where $\textbf{U}$ entails $\textbf{V}$ (for instance, suppose $\textbf{U}$ is $buy$, $\textbf{V}$ is $own$). If we can identify this with predicate entailment detection, we can then answer the question.

To detect predicate entailments, previous work has explored unsupervised learning of typed entailment graphs \cite{szpektor_learning_2008, berant_global_2011,berant_efficient_2015,hosseini_learning_2018,hosseini_duality_2019,hosseini_open-domain_2021}. Entailment graphs are directed graphs, where each node represents the predicate of a relation, and an edge from node $\textbf{U}$ to node $\textbf{V}$ denotes ``$\textbf{U}$ entails $\textbf{V}$''. Entailment graphs are built based on the Distributional Inclusion Hypothesis (DIH) \cite{dagan_similarity-based_1999,geffet_distributional_2005,herbelot_measuring_2013,kartsaklis_distributional_2016}. Predicates are disambiguated according to their arguments' types, predicates taking the same types of arguments go into one subgraph.

While previous work on entailment graphs has mostly been limited to English, building entailment graphs in other languages is interesting and challenging. The importance is two-fold: for that language, a native entailment graph would facilitate NLU in it; for cross-lingual inference, entailment graphs in different languages host exploitable complementary information. In particular, we argue that by jointly consulting strong entailment graphs in multiple languages, improvements can be gained for inference in \textbf{all} participant languages.

In this paper, we choose Chinese as our target language to build entailment graphs, as it is distant enough from English to exhibit rich complementarity, while relatively high-resource.
The main challenge for building Chinese entailment graphs, is to extract reliable \textbf{typed relation triples} from raw corpora as strong input. This involves open relation extraction (ORE) and fine-grained entity typing (FET), which we discuss below.

ORE extracts predicate-argument triples from sentences, where previous work has used rule-based methods over syntactic parsers either directly \cite{fader_identifying_2011,etzioni_open_2011,angeli_leveraging_2015}, or for distant supervision \cite{cui_neural_2018, stanovsky_supervised_2018, kolluru_openie6_2020}. The challenge in ORE can be largely attributed to the poor definition of ``open relations''. The situation worsens in Chinese, as the parts of speech are more ambiguous and many linguistic indicators of relations are poorly represented. 
Previous work on Chinese ORE \cite{qiu_zore_2014,jia_chinese_2018} has defined narrow sets of open relations, failing to identify many relational constructions. Conversely, we propose a novel dependency-based ORE method, which we claim provides comprehensive coverage of relational constructions.

FET assigns types to the arguments of extracted relations, so that word-senses of predicates can be disambiguated. The challenge in Chinese FET lies mainly in the lack of datasets over a suitable type ontology: too coarse a type set would be insufficient for disambiguation, too granular a type set would result in disastrous sparsity in the entailment graph. Following \citet{hosseini_learning_2018}, we use the popular FIGER type set \cite{ling_fine-grained_2012}, and build CFIGER, the first FIGER-labelled Chinese FET dataset via label mapping. Entity typing models built on this dataset show satisfactory accuracy and are helpful for predicate disambiguation.

We evaluate our Chinese entailment graph on the Levy-Holt entailment dataset \cite{levy_annotating_2016, holt_probabilistic_2019} via translation.
Results show that our Chinese entailment graph outperforms baselines by large margins, and is comparable to the English graph. We verify our cross-lingual complementarity hypothesis with an ensemble between English and Chinese graphs, where we show a clear advantage over both monolingual graphs\footnote{This effect remains clear when both monolingual graphs are trained with parallel corpora, verifying that complementarity is behind this gain, not the additional corpus involved. See \S\ref{Sec:EntResults_AblationStudy} for more discussions.}, and set a new SOTA for unsupervised predicate entailment detection.

Our contributions are as follows: 1) we present a novel Chinese ORE method sensitive to a much wider range of relations than previous SOTA, and a Chinese FET dataset, the first under the FIGER type ontology; 2) we construct the first Chinese entailment graph, comparable to its English counterpart; 3) we reveal the cross-lingual complementarity of entailment graphs with an ensemble.\footnote{Our codes and data-sets can be found at \url{https://github.com/Teddy-Li/ChineseEntGraph}}

\section{Background and Related Work}
\label{Sec:RelatedWork}

Predicate entailment detection has been an area of active research. \citet{lin_automatic_1998,weeds_general_2003,szpektor_learning_2008} proposed various count-based entailment scores; \citet{berant_global_2011} proposed to ``globalize'' typed entailment graphs by closing them with transitivity constraints; \citet{hosseini_learning_2018} proposed a more scalable global learning approach with soft transitivity constraints; \citet{hosseini_duality_2019, hosseini_open-domain_2021} further refined the entailment scores with standard and contextual link prediction. 

Our work is closely related to \citet{hosseini_learning_2018}, with key adaptations for Chinese in ORE and FET. Their ORE method is based on a CCG parser \cite{reddy_large-scale_2014}, while ours is based on a dependency parser \cite{zhang_practical_2020}; their FET is done by linking entities to Wikipedia entries,
while we use neural entity typing for the task. 

Dependency parses are less informative than CCG parses, and require heavier adaptation. However, Chinese dependency parsers are currently more reliable than CCG parsers \cite{tse_challenges_2012}. Previous Chinese ORE methods \cite{qiu_zore_2014,jia_chinese_2018} are based on dependency parsers, but they omit many common constructions essential to ORE. In \S\ref{Sec:ChineseORE}, we present the most comprehensive Chinese ORE method so far.

Linking-based entity-typing can be more accurate than neural methods, since the type labels are exact as long as linking is correct. However, current Chinese entity linking methods require either translation \cite{pan_cross-lingual_2019} or search logs \cite{fu_design_2020}. Both hurt linking accuracy, and the latter grows prohibitively expensive with scale.
On the other hand, since the seminal work of \citet{ling_fine-grained_2012}, neural fine-grained entity typing has developed rapidly \cite{yogatama_embedding_2015,shimaoka_neural_2017,chen_hierarchical_2020}, with a common interest in FIGER type set. For Chinese, \citet{lee_chinese_2020} built an ultra-fine-grained entity typing dataset,
 based on which we build our CFIGER dataset via label mapping.
 
\citet{weber_construction_2019} aligned English and German entailment graphs, and showed that the English graph can help with German entailment detection. Yet it was uncertain whether this effect comes from genuine complementarity or the mere fact that the English graph is stronger. We take one step further, and show that complementarity can be exploited in both directions: for English, the higher resource language, entailment detection can also benefit from the ensemble to reach new heights.

As a related resource, \citet{ganitkevitch_ppdb_2013} created a multi-lingual database for symmetric paraphrases; in contrast, entailment graphs are directional. More recently, \citet{schmitt_language_2021} proposed to fine-tune language models on predicate entailment datasets via prompt learning. In contrast to our entailment graphs, their approach is supervised, which carries the danger of overfitting to dataset artifacts \cite{gururangan_annotation_2018}.

Another related strand of research, e.g. SNLI \cite{bowman_large_2015}, is concerned with the more general NLI task, including hypernymy detection and logic reasoning like $A \wedge B \rightarrow B$, but rarely covers the cases requiring external knowledge of predicate entailment \cite{hosseini_learning_2018}. Conversely, entailment graphs aim to serve as a robust resource for directional predicate entailments induced from textual corpora.

\section{Chinese Open Relation Extraction}
\label{Sec:ChineseORE}
We build our ORE method based on DDParser \cite{zhang_practical_2020}, a SOTA Chinese dependency parser. We mine relation triples from its output by identifying patterns in the dependency paths.

Depending on the semantics of the head verb, instances of a dependency pattern can range from being highly felicitous to marginally acceptable as a relation. Motivated by our downstream task of entailment graph construction, we go for higher recall and take them in based on the \textbf{Relation Frequency Assumption}: less felicitous relations occur less frequently, and are less likely to take part in entailments when they do occur, thus they are negligible.

Due to the lack of a commonly accepted benchmark or criterion for ``relations'', we did not perform an intrinsic evaluation for our Chinese ORE method; its significant benefit to our EG$_\textit{Zh}$ graph, as shown in \S\ref{Sec:EntResults}, should suffice to demonstrate its strength.

\subsection{Parsing for Chinese ORE}
\label{Sec:P4CORE}
The task of open relation extraction on top of LM-driven dependency parsers, is really the task of binding the relations in surface forms to the underlying relation structures. Though trivial at first sight, the definition of these underlying and essentially semantic relations demands detailed analysis.

\citet{jia_chinese_2018} is the latest to propose an ORE method on dependency paths. They defined a set of rules to extract relations patterns, which they call dependency semantic normal forms (DSNFs)\footnote{We refer readers to Appendix \ref{supp_jia} for a brief summary.}.

However, their set of DSNFs is inexhaustive and somewhat inaccurate. We show below that many linguistic features of Chinese demand a more principled account, more constructions need to be considered as relations, some to be ruled out. These observations are made from a multi-source news corpus, which we use to build entailment graphs (\S\ref{Sec:CEnt})\footnote{Since entailment graph construction is fully unsupervised, this source corpus is independent of the evaluation in \S\ref{Sec:EvalEnt}. Particularly, the Levy-Holt dataset used in \S\ref{Sec:EvalEnt} consists of short sentences, which is a vastly different genre, involving much simpler structures, with a single relation per sentence and few subordinating constructions discussed above (see Appendix \ref{supp_comparison_webhose_levyholt} for relevant statistics)}.
Below, we highlight 5 additional constructions we identify, explained with examples\footnote{We refer readers to Appendix \ref{supp_diagrams} for diagram illustrations.}. 

\vspace{0.05in}
\noindent\textbf{A. PP Modifiers as ``De'' Structures}\quad
One key feature of Chinese is its prevalent use of \textit{``De''} structures in the place of prepositional phrases, where \textit{``De''} can be seen as roughly equivalent to the possessive clitic \textbf{\textit{'s}}. For instance, in  \textit{``咽炎(pharyngitis) 成为(becomes) 发热(fever) 的(De) 原因(cause); Pharyngitis becomes the cause of fever''}, the 
root clause in Chinese is (Pharyngitis, becomes, cause), but we {\em additionally} extract the underlying relation \textbf{(pharyngitis, becomes·X·De·cause, fever)}, where the true object \textit{``fever''} is a \textbf{nominal} attribute of the direct object \textit{``cause''}, and the true predicate subsumes the direct object\footnote{Here and below, examples are paired with English metaphrases, and when necessary, paraphrases; relation triples are presented as English metaphrases (inflections ignored) \iffalse for readers' convenience \fi.}.

The same also applies to the subject, though somewhat more restricted. For sentences like \textit{``苹果(Apple) 的(De) 创始人(founder) 是(is) 乔布斯(Jobs); The founder of Apple is Jobs''}, we additionally extract the relation \textbf{(Apple, founder·is, Jobs)}, where the true subject \emph{``Apple''} is a \textbf{nominal} attribute of the direct subject \emph{``founder''}, and the true predicate subsumes the direct subject\footnote{These relations are more felicitous with frequent predicate argument combinations, and less so for the infrequent ones. As in line with the Relation Frequency Assumption, less felicitous relations are also less statistically significant.}.

\vspace{0.05in}
\noindent\textbf{B. Bounded Dependencies}\quad
In Chinese, bounded dependencies, especially control structures, are expressed with a covert infinitival marker, equivalent to English ``to''. We capture the following phenomena in addition to direct relations:
\begin{itemize}[itemsep=0ex, topsep=0.5ex, leftmargin=0.3cm]
	\item Sequences of VPs: for sentences like \textit{``我(I) 去(go-to) 诊所(clinic) 打(take) 疫苗(vaccine); I go to the clinic to take the vaccine''}, the two verb phrases ``去(go-to) 诊所(clinic)'' and ``打(take) 疫苗(vaccine)'' are directly concatenated, with no overt connection words. Here we additionally extract the relation \textbf{(I, take, vaccine)} by copying the subject of the head verb to subsequent verbs.
	\item Subject-control verbs: for the famous example \textit{``我(I) 想(want) 试图(try) 开始(begin) 写(write) 一个(a) 剧本(play); I want to try to begin to write a play''}, again the verbs are directly concatenated; this time, all verbs but the first one behaves as infinitival complements to their direct antecedents. In such cases, we extract sequences of relations like \textbf{(I, want, try)}, \textbf{(I, want·try, begin)}, \textbf{(I, want·try·begin, write)}, \textbf{(I, want·begin·try·write, a play)}.
\end{itemize}
Notably, the above relations are different from \citet{jia_chinese_2018}'s conjunction constructions in Table \ref{Tab:jia}: the event sequences here involve subordination (control) rather than coordination, thus require a separate account.

\vspace{0.05in}
\noindent\textbf{C. Relative Clauses}\quad
Relative Clauses also take the form of modification structures in Chinese, for which additional relations should also be extracted. For example, in \textit{``他(he) 解决(solve) 了(-ed) 困扰(puzzle) 大家(everyone) 的(De) 问题(problem); He solved the problem that puzzled everyone''}, we extract not only the direct relation \textbf{(he, solve, problem)}, but also the relation embedded in the modification structure \textbf{(problem, puzzle, everyone)}.

\vspace{0.05in}
\noindent\textbf{D. Nominal Compounds}\quad
Relations can be extracted from nominal compounds, where an NP has two consecutive ``ATT'' modifiers: in \textit{``德国(Germany) 总理(Chancellor) 默克尔(Merkel); German Chancellor Merkel''}, ``Germany'' modifies ``Chancellor'', and ``Chancellor'' modifies ``Merkel''. \citet{jia_chinese_2018} extracted relations like \textbf{(Germany, Chancellor, Merkel)} for these NPs. 

However, they overlooked the fact that prepositional compounds in Chinese with omitted ``De'' take exactly the same form (see construction \textbf{A}). 
For example, in NPs with nested PP modifiers like \textit{``手续(formalities) 办理(handle) 时效(timeliness); Timeliness of the handling of formalities''} 
, we observe the same structure, but it certainly does not mean \textit{``the handling of formalities is timeliness''}!

We take a step back and put restrictions on such constructions: only when all 3 words in the NP are nominals (but not pronouns), the third word is the head, the second is a `PERSON' or `TITLE', and the first is a `PERSON', then it is a relation, like \textbf{(Merkel, is·X·De·Chancellor, Germany)}. Otherwise, such NPs rarely host felicitous relations.

\vspace{0.05in}
\noindent\textbf{E. Copula with Covert Objects}\quad
The copula is sometimes followed by modifiers ending with \textit{``De''}. Examples are \textit{``玉米(Corn) 是(is) 从(from) 美国(US) 引进(introduce) 的(De); Corn is introduced from US''}, \textit{``设备(device) 是(is) 木头(wood) 做(make) 的(De); The device is made of wood''}.

In these cases, there exists an object following the indicator \textit{``的(De)''}, but the object is an empty \textbf{\textit{pro}} considered inferable from context. In the absence of the true object, the \textit{VOB} label is given to \textit{``的(De)''}, leading to direct relations like \textbf{(Corn, is, De)}. However, the true predicates are rather \textit{``is introduced from''} or \textit{``is made of''}. To fix this, we \textbf{replace} the direct relations with ones like \textbf{(Corn, is·from·X·introduce·De·\textit{pro}, America)}, reminiscent of the constructions \textbf{A}.


\subsection{Our ORE Method}
With the above constructions taken into account, we build our ORE method on top of DDParser.
For part-of-speech labels, we use the POS-tagger in Stanford CoreNLP \cite{manning_stanford_2014}.
We detect negations by looking for negation keywords in the adjunct modifiers of predicates:
for predicates with an odd number of negation matches, we insert a negation indicator to them, treating them as separate predicates from the non-negated ones.



\section{Chinese Fine-Grained Entity Typing}
\label{Sec:CFet}

As shown in previous work \cite{berant_global_2011, hosseini_learning_2018}, the types of a predicate's arguments are helpful for disambiguating a predicate in context. To this end, we need a fine-grained entity typing model to classify the arguments into sufficiently discriminative yet populous types.

\citet{lee_chinese_2020} presented CFET dataset, an ultra-fine-grained entity typing dataset in Chinese. They labelled entities in sentence-level context, into around 6,000 free-form types and 10 general types. Unfortunately, their free-form types are too fragmented for predicate disambiguation, and their general types are too ambiguous.

We turn to FIGER \cite{ling_fine-grained_2012}, a commonly used type set: we re-annotate the CFET dataset with FIGER types through label mapping. Given that there are around 6,000 ultra-fine-grained types and only 112 FIGER types (49 in the first layer), we can reasonably assume that each ultra-fine-grained type can be unambiguously mapped to a single FIGER type. For instance, the ultra-fine-grained type ``湖~(lake)'' is unambiguously mapped to the FIGER label ``location / body\_of\_water''.

Based on this assumption, we manually create a mapping between the two, and re-annotate CFET dataset with the mapping. We call the re-annotated dataset \textbf{CFIGER}, as it is the first in Chinese with FIGER labels. As with CFET, this dataset consists of 4.8K crowd-annotated data (equally divided into crowd-train, crowd-dev and crowd-test) and 1.9M distantly supervised data from Wikipedia\footnote{For detailed statistics, please refer to Appendix \ref{supp_cfiger}.}.

For training set we combine the crowd-train and Wikipedia subsets; for dev and test sets we use crowd-dev and crowd-test respectively.
We train two baseline models: \textit{CFET}, the baseline model with CFET dataset; \textit{HierType} \cite{chen_hierarchical_2020}, a SOTA English entity typing model. 

 \begin{table}[t]
	\centering
	\begin{tabular}{|l|c|c|}
		\hline
		Macro F1 (\%) & dev & test \\\hline
		\textit{CFET} with CFET dataset & - & 24.9 \\
		\textit{CFET} with CFIGER dataset & 75.7 & 75.7 \\
		\textit{HierType} with FIGER dataset & - & 82.6 \\
		\textit{HierType} with CFIGER dataset & 74.8 & 74.5 \\\hline
	\end{tabular}
	\caption{F1 scores of baseline models for CFIGER dataset, compared with the results on the datasets where they were proposed. Macro-F1 scores are reported because it is available in both baselines.}
	\label{Tab:fet}
\end{table}

Results are shown in Table \ref{Tab:fet}: 
the F1 score of \textit{HierType} model is slightly lower on CFIGER dataset than on FIGER dataset in English; 
contrarily, thanks to fewer type labels, the F1 score of \textit{CFET} baseline increases on CFIGER, bringing it on par with the more sophisticated \textit{HierType} model. This means our CFIGER dataset is valid for Chinese fine-grained entity typing, and may contribute to a benchmark for cross-lingual entity typing.

For downstream applications, we nevertheless employ the \textit{HierType} model, as empirically it generalizes better to our news corpora. 
As shown in later sections, the resulting FET model can substantially help with predicate disambiguation.

\section{The Chinese Entailment Graph}
\label{Sec:CEnt}
We construct the Chinese entailment graph from the Webhose corpus\footnote{\url{https://webhose.io/free-datasets/chinese-news-articles/}}, a multi-source news corpus of 316K news articles, crawled from 133 news websites in October 2016. Similarly to the NewsSpike corpus used in \citet{hosseini_learning_2018}, the Webhose corpus contains multi-source non-fiction articles from a short period of time. This means it is also rich in reliable and diverse relation triples over a focused set of events, ideal for building entailment graphs.

We cut the articles into sentences by punctuations, limiting the maximum sentence length to 500 characters (the maximum sequence length for Chinese Bert). We discard the sentences shorter than 5 characters, and the articles whose sentences are all shorter than 5 characters. After applying the filter, we are left with 313,718 articles, as shown in Table \ref{Tab:stats}.

For these 314K valid articles in Webhose, we get their CoreNLP POS tags and feed them into our ORE method in \S\ref{Sec:ChineseORE}, to extract the open relation triples. Then, with \textit{HierType} model \cite{chen_hierarchical_2020} in \S\ref{Sec:CFet}, we type all arguments of the extracted relations; we type each predicate with its subject-object type pair, such as \emph{person-event} or \emph{food-law}; following previous work, we consider only the first-layer FIGER types; when multiple type labels are outputted, we consider all combinations as valid types for that predicate.

\begin{table}[t]
	\centering
	\begin{tabular}{|l|c|c|}
		\hline
		 & EG$_\textit{Zh}$ & EG$_\textit{En}$ \\\hline
		\# of articles taken & 313,718 & 546,713 \\\hline
		\# of triples used & 7,621,994 & 10,978,438 \\\hline
		\# of predicates & 363,349 & 326,331 \\\hline\hline
		\multicolumn{3}{|l|}{\# of type pairs where:} \\\hline
		subgraph exists & 942 & 355 \\
		|subgraph| > 100 & 442 & 115 \\
		|subgraph| > 1,000 & 149 & 27 \\
		|subgraph| > 10,000 & 26 & 7 \\
		\hline
	\end{tabular}
	\caption{Stats of our Chinese entailment graph (EG$_\textit{Zh}$) compared with the English graph in \citet{hosseini_learning_2018} (EG$_\textit{En}$). $|\cdot|$ denotes the number of predicates.}
	\label{Tab:stats}
\end{table}

We finally employ the entailment graph construction method in \citet{hosseini_learning_2018}, taking in only binary relations \footnote{We encourage interested readers to also check Appendix \ref{supp_build_entgraph} for a brief introduction to \citet{hosseini_learning_2018}.}.  The detailed statistics of our Chinese entailment graph are shown in Table \ref{Tab:stats}: compared with EG$_\textit{En}$, our graph is built on just over half the number of articles, yet we have extracted 70\% the number of relation triples, and built a graph involving even more predicates. In general, our EG$_\textit{Zh}$ is of comparable size to EG$_\textit{En}$.

We have also considered using another larger corpus, the CLUE corpus, for building the Chinese entailment graphs, but couldn't finish due to limits on computational resources \footnote{Our computing environment is specified in Appendix \ref{supp_ethics}}. The larger corpus is built by \citet{xu_clue_2020}, which is eight times the size of the Webhose corpus and is originally intended for training Chinese language models. We provide the typed relation triples extracted from the CLUE corpus as a part of our release, and encourage interested readers to build their own Chinese entailment graph on this larger corpus, as we expect it to exhibit stronger performance, and present an interesting comparison to the language-model driven models pre-trained with the same corpus.


\section{Evaluation Setup}
\label{Sec:EvalEnt}
\subsection{Benchmark and Baselines}
We evaluate the quality of our Chinese entailment graph with the predicate entailment detection task, on the popular Levy-Holt dataset \cite{levy_annotating_2016, holt_probabilistic_2019}. We use the same dev/test configuration as \citet{hosseini_learning_2018}. We convert the Levy-Holt dataset to Chinese through machine translation, then do evaluation on the translated premise-hypothesis pairs. 

We are painfully aware that translation adds noise; in response, we conduct a human evaluation on 100 entries of Levy-Holt development set, as a proxy to the quality of translation. We find that for 89/100 of the entries, the annotation label remains correct; of which, for 74/100 the entries, the meanings of the translations are precise reflections to the English originals\footnote{For more details please check Appendix \ref{supp_manual_examination}.}. Apart from the human evaluation, we will further discuss the effect of machine-translation in \S\ref{Sec:EntResults}.

In Levy-Holt dataset, the task is: to take as input a pair of relation triples about the same arguments, one premise and one hypothesis, and judge whether the premise entails the hypothesis. For example, given the premise \textit{``John, shopped in, Tesco''}, we would like a model to identify the hypothesis \textit{``John, went to , Tesco''} as being entailed by it.

To convert Levy-Holt dataset into Chinese, we concatenate each relation triple into a pseudo-sentence, use Google Translate to translate the pseudo-sentences into Chinese, then parse them back to Chinese relation triples with our ORE method in \S\ref{Sec:ChineseORE}. If multiple relations are returned, we retrieve the most representative ones, by considering only those relations whose predicate covers the HEAD word.\footnote{See Appendix \ref{supp_select_rel} for more details.}

To type the Chinese relation triples, we again use \textit{HierType} model to collect their subject-object type-pairs. The premise and hypothesis need to take the same types of arguments, so we take the intersection of their possible type-pairs as valid pairs (unless the intersection is empty, in which case we take the union).
We search the entailment subgraphs of these valid type-pairs, for entailment edges from the premise to the hypothesis, and return the entailment scores associated with these edges.
When edges are found from multiple subgraphs, we take their maximum score; when no edge is found with any of the valid type-pairs, we back up to the average score from arbitrary type-pairs.

We compare our Chinese entailment graph with a few strong baselines:

\vspace{0.1in}
\noindent\textbf{\textit{BERT}}: We take the translated premise-hypothesis pairs (as the original pseudo-sentences), and compute the cosine similarity between their pretrained BERT representations at [CLS] token. This is a strong 
baseline but symmetric;

\noindent\textbf{\textit{Jia}}: We build entailment graph in the same way as \S\ref{Sec:CEnt}, but with the baseline ORE method by \citet{jia_chinese_2018}; accordingly, \citet{jia_chinese_2018} method is also used in parsing the translated Levy-Holt pseudo-sentences for this evaluation;

\noindent\textbf{\textit{DDPORE}}: Similar to \textit{Jia} baseline, but with the baseline ORE method from DDParser \shortcite{zhang_practical_2020}.


\subsection{Cross-lingual Ensembles}

In order to examine the complementarity between our Chinese entailment graph (EG$_\textit{Zh}$) and the English graph (EG$_\textit{En}$) \shortcite{hosseini_learning_2018}, we ensemble the predictions from the two graphs, $pred_{en}$ and $pred_{zh}$\footnote{``zh'' is the abbreviation for Chinese by convention.}. We experiment with four ensemble strategies: lexicographic orders from English to Chinese and Chinese to English, max pooling and average pooling:

\begin{align*}
pred_{en\_zh}&=pred_{en} + \gamma*\Theta(pred_{en})*pred_{zh} \\
pred_{zh\_en}&=\gamma*pred_{zh} + \Theta(pred_{zh})*pred_{en} \\
pred_{max}&=\textsc{\textit{MAX}}(pred_{en}, \gamma*pred_{zh}) \\
pred_{avg}&=\textsc{\textit{AVG}}(pred_{en}, \gamma*pred_{zh})
\end{align*}

where $\Theta(\cdot)$ is the boolean function $IsZero$, $\gamma$ is the relative weight of Chinese and English graphs.
$\gamma$ is a hyperparameter tuned on Levy-Holt dev set, searched between 0.0 and 1.0 with step size 0.1.

For instance, suppose our premise is \textit{``he, shopped in, the store''}, and our hypothesis is \textit{``he, went to, the store''}, then our Chinese relations, by translation, would be ``他, 在·X·购物, 商店'' and ``他, 前往, 商店'' respectively. Suppose we find in the English graph an edge from \textit{``shop in''} to \textit{``go to''}, scored $pred\_en=0.6$, and we find in the Chinese graph an edge from ``在·X·购物'' to ``前往'', scored $pred\_zh=0.7$. Then we would have $pred_{en\_zh}=0.6$, $pred_{zh\_en}=0.7$, $pred_{max}=0.7$, $pred_{avg}=0.65$.

In addition to ensembling with EG$_\textit{En}$, we also ensembled our entailment graph with the SOTA English graph EG$_\textit{En}$++ \cite{hosseini_open-domain_2021}. We call the latter ones \textbf{Ensemble++} here and below.

\section{Results and Discussions}
\label{Sec:EntResults}

 \begin{table}[t]
	\centering
	\begin{tabular}{|l|c|c|}
		\hline
		AUC (\%) & dev & test \\\hline
		\textit{BERT} $\star$ & 5.5 & 3.2\\
		\textit{Jia} \shortcite{jia_chinese_2018} $\star$ & 0.9 & 2.4 \\
		\textit{DDPORE} \shortcite{zhang_practical_2020} $\star$ & 9.8 & 5.9 \\
		\textbf{EG$_\textit{Zh}$} $\star$ & \textbf{15.7} & \textbf{9.4}\\\hline
		EG$_\textit{En}$ \shortcite{hosseini_learning_2018} $\diamond$ & 20.7 & 16.5\\
		EG$_\textit{En}$++ \shortcite{hosseini_open-domain_2021} $\diamond$ & 23.3 & 19.5 \\
		\textbf{Ensemble En\_Zh} $\diamond$ & \textbf{28.3} ($\gamma:0.8$) & \textbf{21.2} \\
		\textbf{Ensemble Zh\_En} $\diamond$ & \textbf{27.4} ($\gamma:0.9$) & \textbf{21.5} \\
		\textbf{Ensemble MAX} $\diamond$ & \textbf{29.9} ($\gamma:0.8$) & \textbf{22.1}  \\
		\textbf{Ensemble AVG $\diamond$} & \textbf{30.0} ($\gamma:1.0$) & \textbf{22.1} $\dagger$ \\
		\textbf{Ensemble++ AVG $\diamond$} & \textbf{31.2} ($\gamma:0.3$) & \textbf{24.2 $\dagger$}  \\\hline
		EG$_\textit{Zh}$ \textit{-type} $\star$ & 11.1 & 7.0 \\\hline
		DataConcat En $\diamond$ & 20.6 & 17.8 \\
		DataConcat Zh $\star$ & 19.0 & 14.2 \\
		DataConcat Esb $\diamond$ & 31.8 & 25.0 \\\hline
		BackTrans Esb $\diamond$ & 23.0 & 17.5 \\\hline
	\end{tabular}
	\caption{Area Under Curve values on Levy-Holt dataset, for Chinese entailment graph (EG$_\textit{Zh}$), its baselines, ensembles with English graphs, and ablation studies. EG$_\textit{En}$ is the English graph from \cite{hosseini_learning_2018}; EG$_\textit{En}$++ is the English graph from \cite{hosseini_open-domain_2021}. Entries with $\star$ uses Chinese lemma baseline; entries with $\diamond$ uses English lemma baseline; entries with $\dagger$ are the best ensemble strategies by dev set results.}
	\label{Tab:auc}
 \end{table}

To measure the performance of our constructed Chinese entailment graphs, we follow previous work
in reporting the Precision-Recall (P-R) Curves plotted for successively lower confidence thresholds, and their Area Under Curves (AUC), for the range with $>50\%$ precision.

A language-specific lemma baseline sets the left boundary of recall, by exact match over the lemmatized premise / hypothesis. For our Chinese entailment graph (EG$_\textit{Zh}$) and its baselines, the boundary is set by Chinese lemma baseline. 
For the ensembles, in order to get commensurable AUC values with previous work instead of being over-optimistic, we use the English lemma baseline.

\subsection{Experiment Results}

As shown in Table \ref{Tab:auc}, on the Chinese Levy-Holt dataset, our EG$_\textit{Zh}$ graph substantially outperforms the BERT pre-trained baseline. EG$_\textit{Zh}$ is also far ahead of entailment graphs with baseline ORE methods, proving the superiority of our Chinese ORE method against previous SOTA.

 \begin{figure}[t]
 	\centering
 	\includegraphics[scale=0.42]{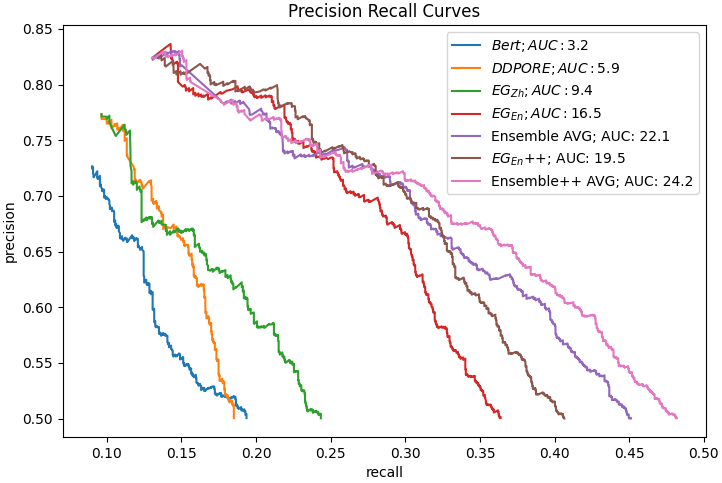}
 	\caption{P-R Curves on Levy-Holt test set for EG$_\textit{Zh}$, ensembles and baselines; \textit{Jia\shortcite{jia_chinese_2018}} baseline is far behind others, thus omitted for the clarity of the figure.}
 	\label{figure_ZhEntGraph}
 \end{figure}

EG$_\textit{Zh}$ and EG$_\textit{En}$ are built with the same algorithm \cite{hosseini_learning_2018}, and evaluated on parallel datasets. Learnt from 57\% the data, EG$_\textit{Zh}$ achieves an AUC exactly 57\% of its English counterpart. Note that the Chinese entailment graph is underestimated with the use of translated dataset: out of the 12,921 relation pairs in Levy-Holt test set, only 9,337 of them are parsed into valid Chinese binary relations. This means, for Chinese entailment graphs, the upper bound for recall is not 100\%, but rather 72.3\%, as is the upper bound for AUC. Besides, the translationese language style in Chinese Levy-Holt also poses a gap in word-choice to the natively-built entailment graph, resulting in more mismatches.
Considering this additional noise, the performance of EG$_\textit{Zh}$ means our pipeline is utilizing information in the source corpus very well.


The ensemble between Chinese and English entailment graphs sets a new SOTA for unsupervised predicate entailment detection. With all 4 ensemble strategies, improvement is gained upon both monolingual graphs; with \textbf{Ensemble AVG}, the best on dev-set, the margin of test set improvement is more than 5 points.
Moreover, when ensembling with EG$_\textit{En}$++, we get a test-set AUC of 24.2 points (\textbf{Ensemble++ AVG}), raising SOTA by 4.7 points.

\subsection{Ablation Studies} 
\label{Sec:EntResults_AblationStudy}

In Table \ref{Tab:auc}, we additionally present three \textbf{ablation studies} to verify the solidarity of our approach.

In the first ablation study, \textit{EG$_\textit{Zh}$ -type}, we take away entity typing and train an untyped entailment graph. In this setting, we lose 2.4 AUC points. This means, our entity typing method, as discussed in \S\ref{Sec:CFet}, is indeed helpful for disambiguating predicates in entailment graphs. 

\begin{table*}[t]
	\centering
	\begin{tabular}{|l|c|c|c|c|c|}
		\hline
		Direct causes of EG$_\textit{Zh}$'s different prediction & TP (+) & FP (-) & TN (+) & FN (-) & +/- \\\hline
		translation-related causes, among which: & +52 & -28 & +42 & -47 & +19 \\
		\quad \textit{· same sentence after translation} & \textit{+52} & \textit{-28} & \textit{0} & \textit{0} & \textit{+24} \\
		\quad \textit{· translation error} & \textit{0} & \textit{0} & \textit{+42} & \textit{-47} & \textit{-5} \\
		lexicalization & +29 & -54 & +16 & -12 & -21 \\
		ORE error & +8 & -20 & +8 & -5 & -9 \\
		\textbf{evidence of entailment} & \textbf{+109} & \textbf{-95} & \textbf{+86} & \textbf{-40} & \textbf{+60} \\\hline
		TOTAL & +198 & -197 & +152 & -104 & +49 \\
		\hline
	\end{tabular}
	\caption{Breakdown of the different predictions between our ensembles and English monolingual graph.  ``TP'', ``FP'', ``TN'', ``FN'' represent \emph{True Positive}, \emph{False Positive}, \emph{True Negative} and \emph{False Negative} respectively; in the column ``+/-'' is the overall impact of each factor.}
	\label{Tab:CaseStudy}
\end{table*}

In the second ablation study, the \textit{DataConcat} settings, we disentangle cross-lingual complementarity from the effect of extra data. We machine-translate NewsSpike corpus into Chinese, and Webhose into English\footnote{ We initially attempted to use Google Translate for translating these larger corpora as well, but turned to Baidu Translate instead for its more generous free quota.}. We build an English graph ``DataConcat En'' using \textit{NewsSpike + translated-Webhose}, and a Chinese graph ``DataConcat Zh'' using \textit{Webhose + translated-NewsSpike}. Results show that while both graphs improve with data from the other side, they are still far behind our \textbf{Ensemble} settings above. Further, we ensembled the two DataConcat graphs as ``DataConcat Esb'', the best dev set setting in this case is MAX ensemble with $\gamma=0.2$. On test set, this ensemble delivers an AUC of 25.0 points, this is 7.2 points higher than DataConcat En, an even wider margin than the non-DataConcat graphs. The above comparison suggests, the success of cross-lingual ensemble \textbf{cannot} be reproduced by sticking all the data together for a monolingual graph.

In the third case study, \textit{BackTrans Esb}, we disentangle cross-lingual complementarity from the effect of machine-translation. Machine translation can be noisy, but it might also map synonyms in the source language to the same words in the target language. To single out this effect, we translate the Chinese Levy-Holt dataset back into English, and perform an ensemble between predictions on the original and the back-translated Levy-Holt. As shown in the last block of Table \ref{Tab:auc}, the gain in this case is only marginal, suggesting that cross-lingual complementarity is the reason for our success, while the synonym effect is not.

In conclusion, from the entailment detection experiment, we have learnt that: 1) our Chinese entailment graph is strong in the monolingual setting, with contributions from the ORE method and entity typing; 2) a cross-lingual complementarity is clearly shown between Chinese and English entailment graphs, where the effect of ensembles is most significant in the moderate precision range (see Figure \ref{figure_ZhEntGraph}). We expect that ensembling strong entailment graphs in more languages would lead to further improvements.

\subsection{Case Study for Cross-lingual Ensembles}
\label{Sec:EntCaseStudy}

Complementary to the discussions above, we further analyse our ensemble with a case study, so as to understand the reasons behind the success of our ensembles against the monolingual graphs. We compare the predictions of our Ensemble\_AVG to that of the English monolingual EG$_\textit{En}$, both thresholded over 65\% precision. We categorize the prediction differences into 4 classes: \textit{True Positives}, \textit{False Positives}, \textit{True Negatives}, \textit{False Negatives}. {\em Positives} are cases where the ensemble switched the prediction label from negative to positive, vice versa for {\em negatives}; {\em True} means that the switch is correct, {\em False}, that the switch is incorrect.

Since the prediction differences between Ensemble\_AVG and EG$_\textit{En}$ is driven by EG$_\textit{Zh}$, in Table \ref{Tab:CaseStudy}, we break down each class of differences according to the direct cause of EG$_\textit{Zh}$ making a different prediction than EG$_\textit{En}$\footnote{examples of each class of cause are given in Appendix \ref{supp_case_study_causes}.}:

\begin{itemize}[itemsep=-0.3ex, topsep=0.3ex, leftmargin=0.32cm]
	\item \textbf{same sentence after translation}: The premise and hypothesis become identical in relation structure; this can only happen with {\em positives};
	
	\item \textbf{translation error}: The premise or hypothesis becomes unparsable into relations due to translation error; this can only happen with {\em negatives};
	
	\item \textbf{lexicalization}: The difference in predictions is attributed to the cross-lingual difference in the lexicalization of complex relations;
	
	\item \textbf{ORE error}: After translation, the true relations in premise and hypothesis have the same arguments, but are mistaken due to ORE error;
	
	\item \textbf{evidence of entailment}: The difference is attributed to the different evidence of entailment in the two graphs; this is most relevant to our EG$_\textit{Zh}$.
\end{itemize}

 As shown, the majority of our performance gain comes from the additional evidence of entailment in EG$_\textit{Zh}$; contrary to intuition, translation played a positive role in the ensemble, though not a major contributor.
 We attribute this to the fact that MT systems tend to translate semantically similar sentences to the same target sentence, though this similarity is still symmetric, not directional. We have singled out this effect in the third ablation study above, and have confirmed that this effect is marginal to our success.

 In Table \ref{Tab:CaseStudy}, for both the differences from evidence of entailment, and differences in TOTAL, the precision of {\em positives} is lower than that of {\em negatives}. Namely, $TP/(TP+FP)$ is lower than $TN/(TN+FN)$. This is no surprise, as {\em positives} and {\em negatives} have different baselines to start with: {\em Positives} attempt to correct the false negatives from EG$_{\it En}$, where 17\% of all negatives are false; {\em Negatives} attempt to correct the false positives, where 35\% of all positives are false (as dictated in the setting of our case study).
 In this context, it is expectable that our evidence of entailment gets $109/(109+95)=53\%$ correct for {\em positives}, while a much better $86/(86+40)=68\%$ correct for {\em negatives}. These results support the solidarity of our contributions.

\section{Conclusion}
\label{Sec:Conclusion}
We have presented a pipeline for building Chinese entailment graphs. Along the way, we proposed a novel high-recall open relation extraction method, and built a fine-grained entity-typing dataset via label mapping. As our main result, we have shown that: our Chinese entailment graph is comparable with English graphs, where unsupervised BERT baseline did poorly; an ensemble between Chinese and English entailment graphs substantially outperforms both monolingual graphs, and sets a new SOTA for unsupervised entailment detection. 
Directions for future work include multilingual entailment graph alignment and alternative approaches for predicate disambiguation.

\section*{Acknowledgements}
The authors would like to thank Jeff Pan for helpful discussions and the anonymous reviewers for their valuable feedback. This work was supported partly by ERC Advanced Fellowship GA 742137 SEMANTAX, a Mozilla PhD scholarship at Informatics Graduate School and the University of Edinburgh Huawei Laboratory.

\bibliography{personal}
\bibliographystyle{acl_natbib}
\newpage
\appendix
\section{A Brief Summary of \citet{jia_chinese_2018}}
\label{supp_jia}
In Table \ref{Tab:jia} are the 7 rules from \citet{jia_chinese_2018} which they call Dependency Structure Normal Forms. The first rule corresponds to nominal compounds which we elaborated in constructions \textbf{D} in \S\ref{Sec:P4CORE}; the second rule corresponds to direct S-V-O relations; the third rule attends to the semantic objects hidden in adjuncts, which are always pre-verbs in Chinese; the fourth rule subsumes complements of head verbs into the predicate; the fifth rule handles the coordination of subjects, the sixth handles coordination of object, and the seventh handles coordination of predicates. These rules are reflected in our ORE method as well, but for the sake of brevity, only the constructions which have never been covered by previous work are listed in \S\ref{Sec:P4CORE}.

\begin{table}[h]
	\centering
	\begin{tabular}{|c|}
		\hline
		德国\quad 总理\quad 默克尔\quad。\\
		German Chancellor Merkel .\\
		(German, Chancellor, Merkel) \\\hline
		我\quad 看到\quad 你\quad。\\
        I see you .\\
        (I, see, you)\\\hline
        他\quad 在\quad 家\quad 玩\quad 游戏\quad。\\
        He at home play game .\\
        (He, play-game, home)\\\hline
        我\quad 走\quad 到\quad 图书馆\quad。\\
        I walk to library .\\
        (I, walk-to, library)\\\hline
        我\quad 和\quad 你\quad 去\quad 商店\quad 。\\
        I and you go-to shop .\\
        (I, go-to, shop)\quad (you, go-to, shop)\\\hline
        我\quad 吃\quad 汉堡\quad 和\quad 薯条\quad。\\
        I eat burger and chips .\\
        (I, eat, burger)\quad (I, eat, chips)\\\hline
        罪犯\quad 击中\quad 、\quad 杀死\quad 了\quad 他\quad。\\
        Criminal shot, kill -ed him .\\
        (criminal, shot, him)\quad (criminal, kill, him)\\\hline
	\end{tabular}
	\caption{Set of DSNFs from \citet{jia_chinese_2018} exemplified. In each box, at top is an example sentence, presented in Chinese and its English metaphrase (inflection ignored); below are the relations they extract.}
	\label{Tab:jia}
\end{table}

\section{Detailed Statistics of the CFIGER dataset}
\label{supp_cfiger}
To test our assumption that each ultra-fine-grained type can be unambiguously mapped to a single FIGER type, we inspect the number of FIGER type labels to which each ultra-fine-grained type is mapped through manual labelling. Among the 6273 ultra-fine-grained types in total, 5622 of them are mapped to exactly one FIGER type, another 510 are not mapped to any FIGER types; only 134 ultra-fine-grained types are mapped to 2 FIGER types, and 7 mapped to 3 FIGER types. No ultra-fine-grained types are mapped to more than 3 FIGER types. Therefore, it is safe to say that our no-ambiguity assumption roughly holds.

We further inspected the number of FIGER types each mention is attached with. It turns out that among the 1,913,197 mentions in total, 59,517 of them are mapped to no FIGER types, 1,675,089 of them are mapped to 1 FIGER type, 160,097 are mapped to 2 FIGER types, 16,309 are mapped to 3 FIGER types, 1,952 are mapped to 4 FIGER types, 200 are mapped to 5 FIGER types, and 33 are mapped to 6 FIGER types. No mentions are mapped to more than 6 FIGER types. Note that each mention can be mapped to more than one ultra-fine-grained types from the start, so these numbers are not in contradiction with the above numbers.

 \begin{figure}[h]
 	\centering
 	\includegraphics[scale=0.37]{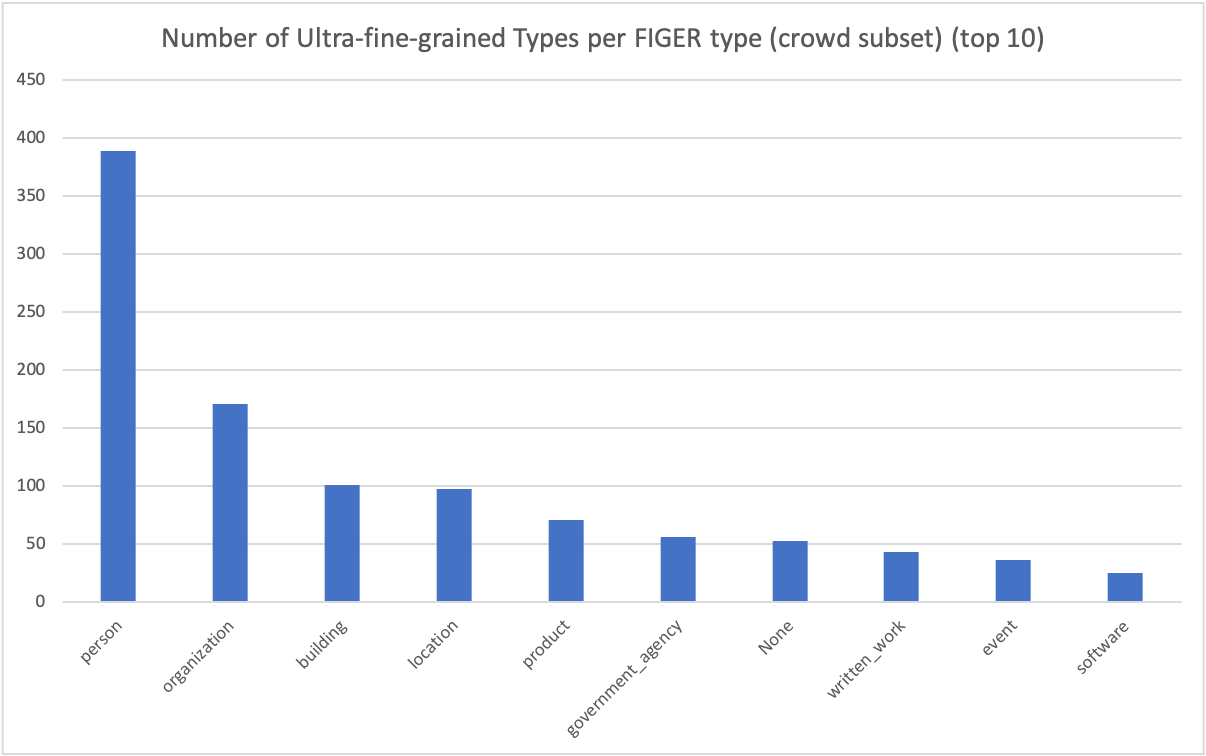}
 	\caption{Number of ultra-fine-grained types in crowd-annotated subset mapped to each FIGER type; only the FIGER types with top 10 number of ultra-fine-grained types are displayed.}
 	\label{figure_crowd_f2u}
 \end{figure}
 
  \begin{figure}[h]
 	\centering
 	\includegraphics[scale=0.37]{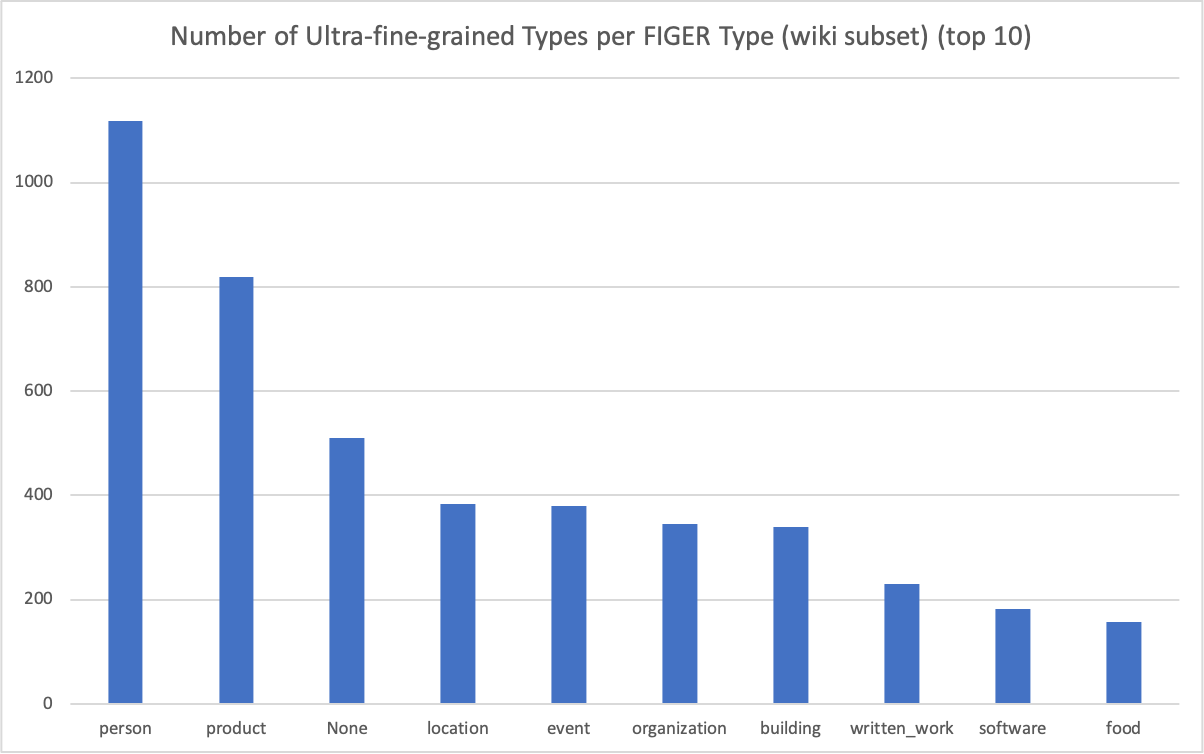}
 	\caption{Number of ultra-fine-grained types in wikipedia distantly supervised subset mapped to each FIGER type; only the FIGER types with top 10 number of ultra-fine-grained types are displayed.}
 	\label{figure_wiki_f2u}
 \end{figure}

We also looked at the number of ultra-fine-grained types each FIGER type is mapped to, so as to understand the skewness of our mapping. Results are shown in Figure \ref{figure_crowd_f2u} and \ref{figure_wiki_f2u}. Unsurprisingly, the most popular ultra-fine-grained labels are highly correlated with the ones that tend to appear in coarse-grained type sets, with ``PERSON'' label taking up a large portion. This distribution is largely consistent between crowd-annotated and Wikipedia subsets.

\begin{figure}[htp]
 	\centering
 	\includegraphics[scale=0.37]{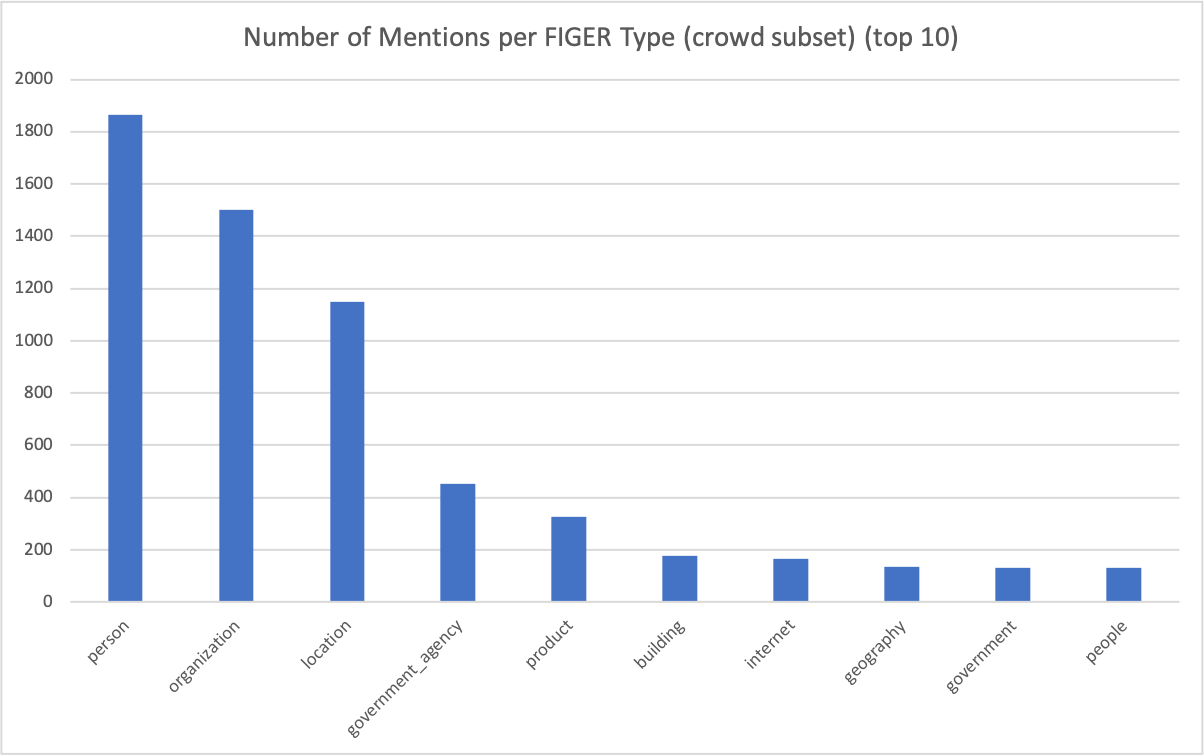}
 	\caption{Number of mentions in crowd-annotated subset labelled as each FIGER type; only the FIGER types with top 10 number of mentions are displayed.}
 	\label{figure_crowd_f2m}
 \end{figure}

\begin{figure}[htp]
 	\centering
 	\includegraphics[scale=0.37]{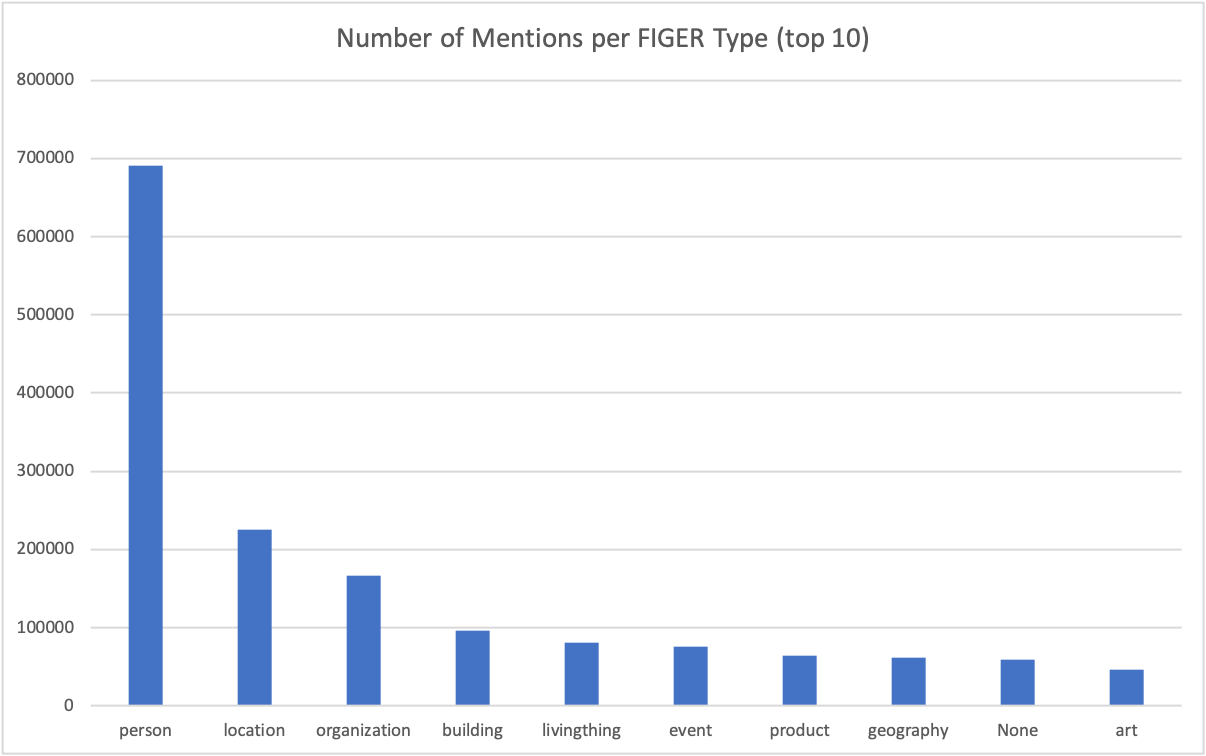}
 	\caption{Number of mentions in wikipedia distantly supervised subset labelled as each FIGER type; only the FIGER types with top 10 number of mentions are displayed.}
 	\label{figure_wiki_f2m}
 \end{figure}

Another set of stats are the number of mentions that corresponds to each FIGER type, shown in Figure \ref{figure_crowd_f2m} and \ref{figure_wiki_f2m}. The winners in terms of the number of mentions are consistent with that of the number of ultra-fine-grained types, and also consistent among themselves (between the two subsets).

\section{Selecting Relation Triples for Translated Levy-Holt}
\label{supp_select_rel}
To retrieve the relation triple most likely reflecting the meaning of the whole sentence, we follow this order when determining which relation triple to select:
\begin{itemize}
    \item For the amended relations, if the predicate of any of them cover the word with HEAD token in DDParser dependency parse, we randomly choose one of these;
    \item If none is found, but the predicate of any non-amended relations cover the word with HEAD token in DDParser dependency parse, we randomly choose one of these;
    \item If none is found, but there are any other relations, we randomly choose one of these;
    \item Finally, if still none is found, we assign \textsc{premise\_placeholder} to the premise and \textsc{hypothesis\_placeholder} to the hypothesis, so that no entailment relation would ever be detected between them.
\end{itemize}

\section{Implementation Details for Entailment Graph Construction}
\label{supp_build_entgraph}

We have used the same entailment graph construction algorithm as \citet{hosseini_learning_2018} to build our Chinese entailment graph from the pool of typed relation triples. When building our entailment graphs, we only feed in the relation triples whose predicate and arguments both appear at least 2 times\footnote{We experimented with 2-2, 2-3, 3-2 and 3-3, among which this 2-2 setting is empirically favoured.}. Their approach of building entailment graphs comes in two steps, in the paragraphs below we will briefly summarize each step and discuss our implementation details.

The first step is local learning. In this step, instances of relation triples are grouped into clusters based on the arguments they take. Relations (predicates) that are seen with the same arguments of the same types are considered to have co-occurred. For each pair of predicates, based on the co-occurrence information, a few different entailment scores have been proposed, of which the BInc score \cite{szpektor_learning_2008} was found to have the best empirical performance in \cite{hosseini_learning_2018}. Following them, we also use the BInc score in the local learning step of our Chinese entailment graphs. Note that after the local learning step, the entailment scores between each pair of predicates are independently calculated, and there are no interactions between entailment subgraphs of different type pairs, thus the name \textbf{local} learning.

The second step is global learning. In this step, global transitivity constraint is ``softly'' applied to the local graphs as an optimization problem: paraphrase predicates are encouraged to have the same pattern of entailment; different typed subgraphs are encouraged to have the same entailment score for the same (ignoring type) pair of predicates; finally, the global scores are encouraged to stay similar to the local scores as a measure of regularization. In \textit{Jia} baseline, the local graphs are too weak for global learning to be helpful; in \textit{DDPORE} baseline, the best dev set AUC (as reported in Table \ref{Tab:auc}) is achieved after 2 epochs; in EG$_\textit{Zh}$, the best dev set AUC is achieved after 3 epochs.

\section{Examples of Different Predictions in Case Study by Category of Direct Cause}
\label{supp_case_study_causes}
In this section, we provide one example for each class of direct cause, as described in \S\ref{Sec:EntCaseStudy}. Chinese sentences and relations in the examples are presented in the same format as \S\ref{Sec:P4CORE}.

\paragraph{Same sentence after translation}
\begin{itemize}[itemsep=0ex, topsep=0.5ex, leftmargin=0.3cm]
    \item Premise - English: (magnesium sulfate, relieves, headache)
    \item Hypothesis - English: (magnesium sulfate, alleviates, headaches)
    \item Premise - Chinese translation: ``硫酸镁(magnesium) 缓解(relieves) 头痛(headache)''
    \item Hypothesis - Chinese translation: ``硫酸镁(magnesium) 缓解(alleviates) 头痛(headache)''
\end{itemize}

The two sentences are translated to the same surface form in Chinese, as the predicates are in many cases synonyms. There are more true positives than false positives, because synonyms are simultaneously more likely true entailments and more likely translated to the same Chinese word.

\paragraph{Translation Error}
\begin{itemize}[itemsep=0ex, topsep=0.5ex, leftmargin=0.3cm]
    \item Premise - English: (Refuge, was attacked by, terrorists)
    \item Hypothesis - English: (Terrorists, take, refuge)
    \item Premise - Chinese translation: ``避难所(refuge) 遭到(suffered) 恐怖分子(terrorists) 袭击(attack); Refuge suffered attack from terrorists.''
    \item Hypothesis - Chinese translation: ``恐怖分子(terrorists) 避难(take-shelter); Terrorists take shelter.''
\end{itemize}

The hypothesis is supposed to mean ``The terrorists took over the refuge''. However, with translation, the hypothesis in Chinese is mistaken as an intransitive relation where take-refuge is considered a predicate.

\paragraph{Lexicalization}
\begin{itemize}[itemsep=0ex, topsep=0.5ex, leftmargin=0.3cm]
    \item Premise - English: (Granada, is located near, mountains)
    \item Hypothesis - English: (Granada, lies at the foot of, mountains)
    \item Premise - Chinese translation: ``格拉纳达(Granada) 靠近(is-near) 山脉(mountains)''
    \item Hypothesis - Chinese translation: ``格拉纳达(Granada) 位于(is-located-at) 山脚下(hillfoot)''
\end{itemize}

When the hypothesis is translated into Chinese, the lexicalization of the relation changed, the part of the predicate hosting the meaning of 'the foot of' is absorbed into the object. Therefore, while in English ``is located near'' does not entail ``lies at the foot of'', in Chinese ``is-near'' is considered to entail ``is-located-at''. In this way, an instance of {\em false positive} comes into being.

\paragraph{ORE Error}
\begin{itemize}[itemsep=0ex, topsep=0.5ex, leftmargin=0.3cm]
    \item Premise - English: (A crow, can eat, a fish)
    \item Hypothesis - English: (A crow, feeds on, fish)
    \item Premise - Chinese translation: ``乌鸦(crow) 可以(can) 吃(eat) 鱼(fish)''
    \item Hypothesis - Chinese translation: ``乌鸦(crow) 以(take) 鱼(fish) 为(as) 食(food)''
    \item Premise - extracted Chinese relation: (crow, eat, fish)
    \item Hypothesis - extracted Chinese relation: (crow, take·X·as·food, fish)
\end{itemize}

While the translations for this pair of relations is correct, in the subsequent Chinese open relation extraction, our ORE method failed to recognize ``可以(can)'' as an important part of the predicate. To avoid sparsity, most adjuncts of the head verb are discarded, and modals are part of them. While the original premise ``can eat'' does not entail ``feeds on'', the Chinese premise ``eat'' does in a way entail ``feeds on'', where another instance of {\em false positive} arises.

\paragraph{Evidence of Entailment}
\begin{itemize}[itemsep=0ex, topsep=0.5ex, leftmargin=0.3cm]
    \item Premise - English: (quinine, cures, malaria)
    \item Hypothesis - English: (quinine, is used for the treatment of, malaria)
    \item Premise - Chinese translation: ``奎宁(quinine) 治疗(cure) 疟疾(malaria)''
    \item Hypothesis - Chinese translation: ``奎宁(quinine) 用于(is-used-to) 治疗(cure) 疟疾(malaria)''
    \item Premise - extracted Chinese relation: (quinine, cure, malaria)
    \item Hypothesis - extracted Chinese relation: (quinine, is-used-to·cure, malaria)
\end{itemize}

In the above example, sufficiently strong evidence for ``cure'' entailing ``is used for the treatment of'' is not found in the English graph, whereas strong evidence for ``治疗(cure)'' entailing ``用于·治疗(is-used-to·cure)'' is found in the Chinese graph. In this way we get an instance of {\em true positive}.

\section{More Precision-Recall Curves}

\begin{figure}[t]
 \centering
 \includegraphics[scale=0.48]{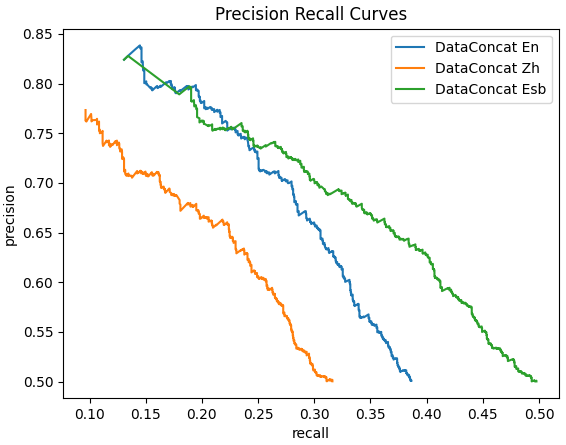}
 \caption{P-R Curves on Levy-Holt test set for DataConcat ablation study.}
 \label{figure_DataConcat}
\end{figure}

\begin{figure}[t]
 \centering
 \includegraphics[scale=0.48]{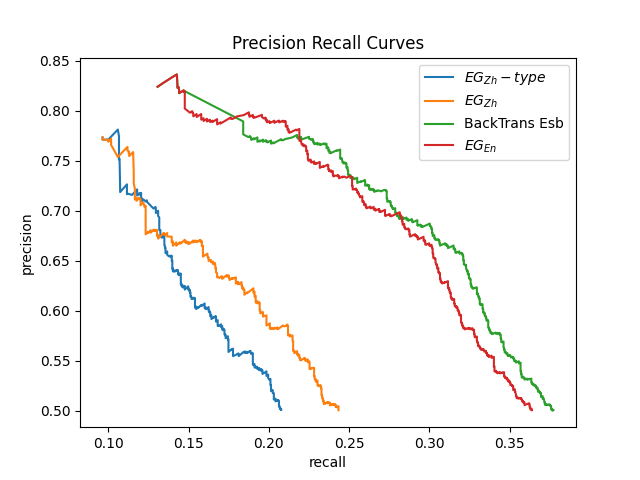}
 \caption{P-R Curves on Levy-Holt test set for EG$_\textit{Zh}$ $-type$, BackTrans Esb, in comparison to EG$_\textit{Zh}$ and EG$_\textit{En}$ respectively.}
 \label{figure_OtherAblations}
\end{figure}

In this section, we present more precision-recall curves from the baselines and ablation studies in Table \ref{Tab:auc}. These curves contain more details explaining the AUC values in the table.

Figure \ref{figure_DataConcat} contains the curves for the ablation study of DataConcat. Here all three models ultimately come from the same corpus, so the performance difference can be fully attributed to the cross-lingual complementarity of entailment graphs.

Figure \ref{figure_OtherAblations} contains the curves for two sets of ablation studies: EG$_\textit{Zh}$ with or without entity typing; EG$_\textit{En}$ ensembled with back-translation predictions or not. The former study shows the clear benefit of our entity typing system, while the latter study shows that ensembling with back-translated predictions only results in a marginal gain, therefore the synonym effect from translation is not a major contributor to the success of our ensembling method.

\section{Manual Examination of Chinese Levy-Holt}
\label{supp_manual_examination}

In order to provide a quantified evaluation for the quality of our Chinese Levy-Holt dataset from a human perspective, we manually labelled 100 proposition pairs in the Chinese Levy-Holt dev set (1-29, 1124-1136, 2031-2059, 3091-3122, 4061-4089, excluding the entries which are not parsed back into binary relation triples).

In this evaluation, we aim to answer the question of ``how accurate is the translate-then-parse procedure when it claims to have successfully converted an evaluation entry''. We label each Chinese entry along two dimensions: semantic consistency, whether it has preserved the meaning of the English entry; label consistency, whether the entailment label remains correct.

Along the first dimension of semantic consistency, we summarize our findings as follows:

\begin{itemize}[itemsep=0ex, topsep=0.5ex, leftmargin=0.3cm]
    \item Correct: 74/100. These are the Chinese entries whose Chinese \textbf{predicates} precisely reflects the meaning of the English entry\footnote{Arguments are allowed to be translated to different senses of the words, as long as the entailment label between the predicates is not affected.};
    \item Metaphors: 3/100. These are the cases where the English entry involves metaphorical word-senses of predicates, but such metaphorical senses of these words are infelicitous in Chinese context;
    \item Adjuncts: 9/100. These are the cases where a part of an English predicate is translated into an adjunct to the Chinese head-verb, and is not included in the Chinese predicate (as in the example for ORE Errors in Appendix \ref{supp_case_study_causes}); examples of missed-out adjuncts are `widely', `should' and `may';
    \item Lexical: 5/100. These are the cases where the word-segmentation of the Chinese sentence is incorrect (as Chinese sentences come with no spaces between words);
    \item Errors: 7/100. These are the cases where, although the Chinese ORE method outputs some binary relation triples for the translation, that relation triple is not the true relation for the sentence;
    \item Translation: 2/100. These are the cases where, although the translation can be parsed into some binary relation triples by our Chinese ORE method, the translation is incorrect, thus everything downstream is wrong.
\end{itemize}

Along the second dimension of label consistency, we find that: in 89 / 100 entries, the actual labels in Chinese are consistent with the English labels; in 10 / 100 entries, the actual labels in Chinese are inconsistent with the English labels; in the remaining 1 / 100 entry, the actual label in Chinese is consistent with the actual label in English, but the provided English label is corrupted.

In summary, for the portion where the conversion is successful, the entries in Chinese Levy-Holt preserves the meaning of the English entries reasonably well; more importantly, the labels of the Chinese Levy-Holt dataset remains robust.

\section{Diagram Illustrations of Our Syntactic Analysis}
\label{supp_diagrams}

In this section, we present for interested readers a set of diagram illustrations of the set of constructions, as involved in our syntactic analysis in \S\ref{Sec:P4CORE}. For each construction, we draw a diagram to illustrate its dependency structure, an example to instantiate the dependency structure, and in the following lines, all the relations that we extract from this construction (one relation per line). Each relation comes in the form of triple-of-types (consistent with the diagram) and triple-of-words (as in the example), separated by semi-colons. The diagrams are presented in Table \ref{Tab:OREDiagrams1}, Table \ref{Tab:OREDiagrams2} and Table \ref{Tab:OREDiagrams3}.

\section{Ethics Considerations}
\label{supp_ethics}
Below we discuss the ethics considerations in our work.

The limitation to our work is two-fold. Firstly, our Chinese entailment graphs focus on the task of predicate entailment detection, and does not attempt to independently solve the more general problem of reasoning and inference: this more general task would also involve other resources including argument hypernymy detection, quantifier identification and co-reference resolution. These are out of the scope of this work. Secondly, while we have shown the effect of cross-lingual complementarity, adding in more languages to the ensemble is not directly straight-forward: this would require linguistic expertise and NLP infrastructure in the respective languages; including more languages, and eventually including arbitrary languages, is one of the directions for our future work.

The risk of our work mostly stems from our use of large-scale news corpora: if the media coverage itself is biased toward certain aspects of the world or certain groups of people, then these biases would be inherited by our entailment graphs. Our response to this is to include as many diverse news sources as possible to reduce such biases to the minimum:  our source corpus for building Chinese entailment graphs includes 133 different news sources from a variety of countries and regions.

For the computational cost of building Chinese entailment graphs, the algorithm for open relation extraction takes roughly 140 CPU hours to process the entirety of Webhose corpus; the entity typing model takes roughly 180 GPU hours on NVidia 1080Ti GPUs to do inference on the entirety of Webhose corpus; the local learning process takes less than one hour, and, the global learning process, our major computational bottleneck, takes roughly 800 CPU hours to finish.

The major datasets of use, namely, Webhose corpus, CLUE dataset and the CFET dataset, are open corpora with no specified licenses, thus our academic use is allowed; no license was specified for the Levy-Holt dataset as well; our own CFIGER dataset as well as the constructed entailment graphs can be distributed under the MIT license.

\section{Comparison Between Webhose Corpus and Levy-Holt Dataset}
\label{supp_comparison_webhose_levyholt}

In this section, we report some key statistics of the Webhose corpus in comparison to the Levy-Holt dataset, which highlight their difference in genre. 

 \begin{table}[t]
	\centering
	\begin{tabular}{|m{3.3cm}|c|c|}
		\hline
		Stats & Webhose & Levy-Holt \\\hline
		AVG sentence length (in \# of Chinese characters) & 24.9 & 10.1\\\hline
		AVG \# of relations per sentence & 15.6 & 2.72 \\\hline
		Percentage of relations from our additional patterns in 
		\S\ref{Sec:P4CORE} & 48\% & 32\% \\\hline
	\end{tabular}
	\caption{Some key statistics of Webhose corpus and Chinese Levy-Holt dataset.}
	\label{Tab:corpusStats}
 \end{table}

As shown in Table \ref{Tab:corpusStats}, the Webhose corpus has much longer sentences than the Chinese Levy-Holt dataset, and on average, a much larger number of open relations can be extracted from the sentences in Webhose corpus. More importantly, the relation patterns which we additionally identified in \S\ref{Sec:P4CORE} are much better represented (constituting 48\% of all relations) than in Chinese Levy-Holt (32\%). Thus, it is clear that: 1) our ORE method in \S\ref{Sec:ChineseORE} was not tuned on the test data, namely Chinese Levy-Holt; 2) tuning on Chinese Levy-Holt would not help with building better ORE methods for news corpora. On the other hand, as a large-scale multi-source news corpus of 5 million sentences, Webhose corpus can be believed to accurately reflect the distribution of linguistic patterns in the entirety of the news genre.

\begin{table*}[t]
  \centering
  \renewcommand{\arraystretch}{1.5}
  \begin{tabular}{ | m{2.5cm} | m{12.6cm} | }
    \hline
    Construction ID & Diagrams and Examples \\ [0.1cm] \hline
    \multirow{4}{*}{\textbf{A.1}} &
    \begin{minipage}{1\textwidth}
      \includegraphics[scale=0.2]{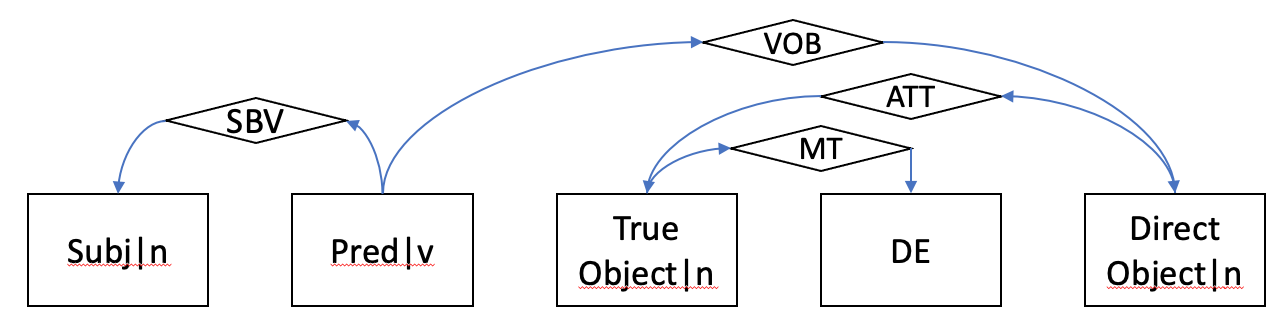}
    \end{minipage}
    \\\cline{2-2}
    & 
    \textbf{Example}: \textit{``咽炎(pharyngitis) 成为(becomes) 发热(fever) 的(De) 原因(cause); Pharyngitis becomes the cause of fever''}
    \\\cline{2-2}
    &
    \textbf{Relation 1}: (Subj, Pred, Direct\_Object);\quad (咽炎(pharyngitis), 成为(becomes), 原因(cause))
    \\\cline{2-2}
    &
    \textbf{Relation 2}: (Subj, Pred·X·DE·Direct\_Object, True\_Object);\quad (咽炎(pharyngitis), 成为·X·的·原因(becomes·X·DE·cause), 发烧(fever))
    \\\hline
    
    \multirow{4}{*}{\textbf{A.2}} &
    \begin{minipage}{.3\textwidth}
      \includegraphics[scale=0.22]{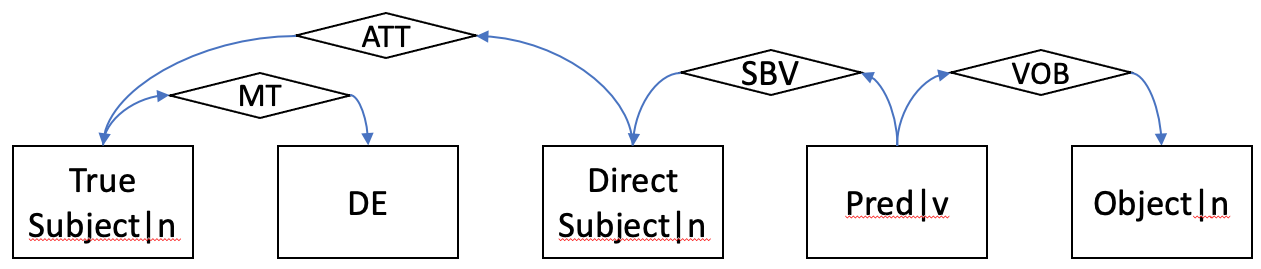}
    \end{minipage}
    \\\cline{2-2}
    & 
    \textbf{Example}: \textit{``苹果(Apple) 的(De) 创始人(founder) 是(is) 乔布斯(Jobs); The founder of Apple is Jobs''}
    \\\cline{2-2}
    &
    \textbf{Relation 1}: (Direct\_Subject, Pred, Object);\quad (创始人(founder), 是(is), 乔布斯(Jobs))
    \\\cline{2-2}
    &
    \textbf{Relation 2}: (True\_Subject, Direct\_Subject·Pred, Object);\quad (苹果(Apple), 创始人·是(founder·is), 乔布斯(Jobs))
    \\\hline
    
    \multirow{4}{*}{\textbf{B.1}} &
    \begin{minipage}{.3\textwidth}
      \includegraphics[scale=0.22]{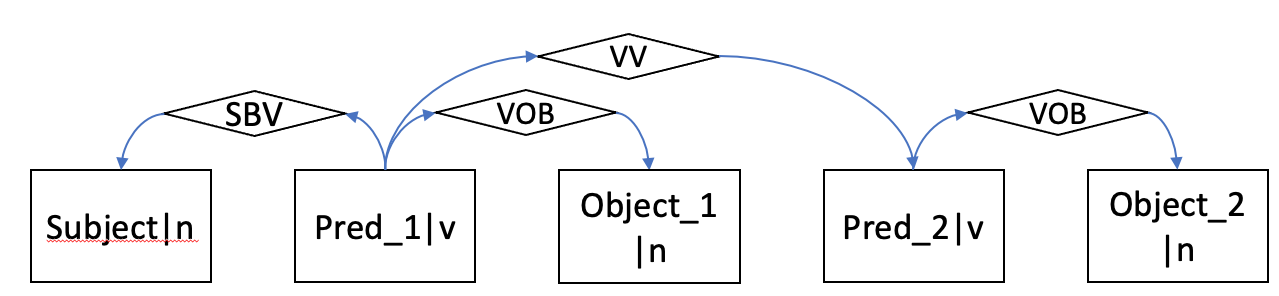}
    \end{minipage}
    \\\cline{2-2}
    & 
    \textbf{Example}: \textit{``我(I) 去(go-to) 诊所(clinic) 打(take) 疫苗(vaccine); I go to the clinic to take the vaccine''}
    \\\cline{2-2}
    &
    \textbf{Relation 1}: (Subject, Pred\_1, Object\_1);\quad (我(I), 去(go-to), 诊所(clinic))
    \\\cline{2-2}
    &
    \textbf{Relation 2}: (Subject, Pred\_2, Object\_2);\quad (我(I), 打(take), 疫苗(vaccine))
    \\\hline
    
  \end{tabular}
  \caption{The syntactic analysis in \S\ref{Sec:P4CORE} illustrated with diagrams, examples and their extracted relations.}\label{Tab:OREDiagrams1}
\end{table*}

\begin{table*}[t]
  \centering
  \renewcommand{\arraystretch}{1.5}
  \begin{tabular}{ | m{2.5cm} | m{12.6cm} | }
    \hline
    Construction ID & Diagrams and Examples \\ [0.1cm] \hline
    
    \multirow{6}{*}{\textbf{B.2}} &
    \begin{minipage}{.3\textwidth}
      \includegraphics[scale=0.22]{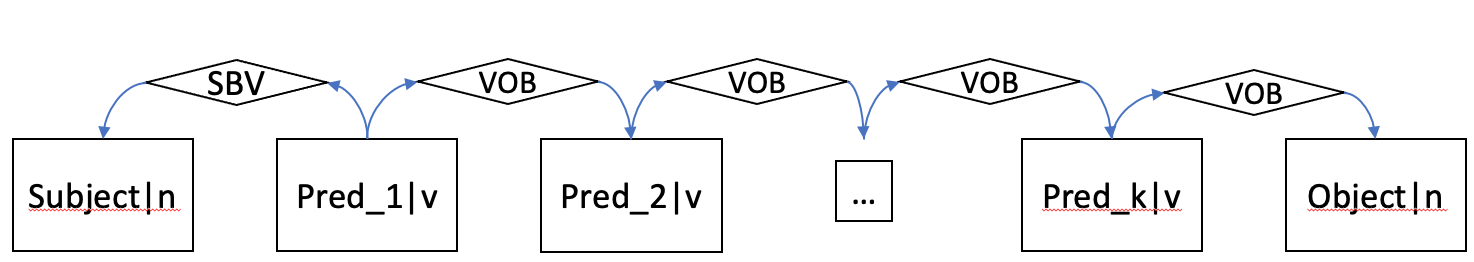}
    \end{minipage}
    \\\cline{2-2}
    & 
    \textbf{Example}: \textit{``我(I) 想(want) 试图(try) 开始(begin) 写(write) 一个(a) 剧本(play); I want to try to begin to write a play''}
    \\\cline{2-2}
    &
    \textbf{Relation 1}: (Subject, Pred\_1, Pred\_2);\quad (我(I), 想(want-to), 试图(try))
    \\\cline{2-2}
    &
    \textbf{Relation 2}: (Subject, Pred\_1·Pred\_2, Pred\_3);\quad (我(I), 想·试图(want-to·try), 开始(begin))
    \\\cline{2-2}
    &
    ......
    \\\cline{2-2}
    &
    \textbf{Relation K}: (Subject, Pred\_1·...·Pred\_K, Object);\quad (我(I), 想·试图·开始·写(want-to·try·begin·write), 一个剧本(A play))
    \\\hline
    
    \multirow{4}{*}{\textbf{C}} &
    \begin{minipage}{.3\textwidth}
      \includegraphics[scale=0.22]{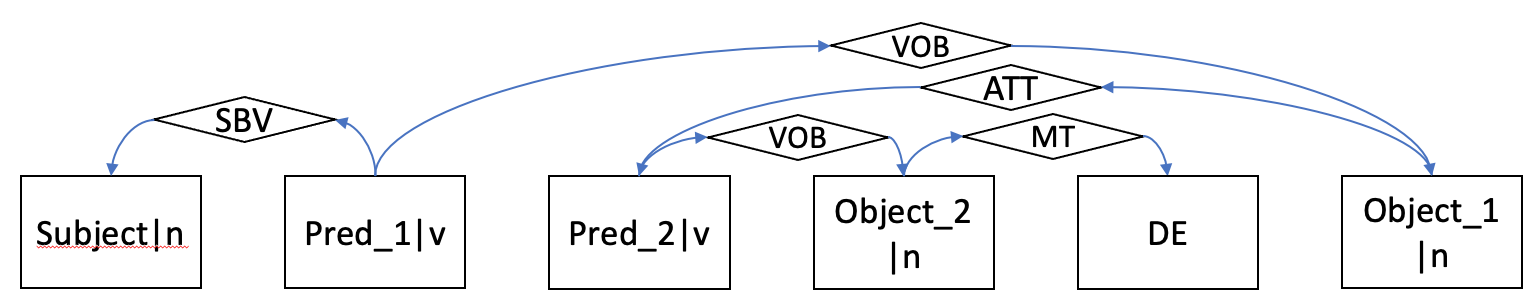}
    \end{minipage}
    \\\cline{2-2}
    & 
    \textbf{Example}: \textit{``他(he) 解决(solve) 了(-ed) 困扰(puzzle) 大家(everyone) 的(De) 问题(problem); He solved the problem that puzzled everyone''}
    \\\cline{2-2}
    &
    \textbf{Relation 1}: (Subject, Pred\_1, Object\_1);\quad (他(He), 解决(solved), 问题(problem))
    \\\cline{2-2}
    &
    \textbf{Relation 2}: (Object\_1, Pred\_2, Object\_2);\quad (问题(Problem), 困扰(puzzled), 大家(everyone))
    \\\hline
    
    \textbf{D} & Analysis in construction D removes the infelicitous instances of the \textbf{Nominal Compound} construction; for the illustration of this construction, we refer readers to \citet{jia_chinese_2018} and do not repeat here. \\\hline
    
  \end{tabular}
  \caption{More syntactic analysis in \S\ref{Sec:P4CORE} illustrated with diagrams, examples and their extracted relations.}\label{Tab:OREDiagrams2}
\end{table*}

\begin{table*}[t]
  \centering
  \renewcommand{\arraystretch}{1.5}
  \begin{tabular}{ | m{2.5cm} | m{12.6cm} | }
    \hline
    Construction ID & Diagrams and Examples \\ [0.1cm] \hline
    
    \multirow{4}{*}{\textbf{E.1}} &
    \begin{minipage}{.3\textwidth}
      \includegraphics[scale=0.22]{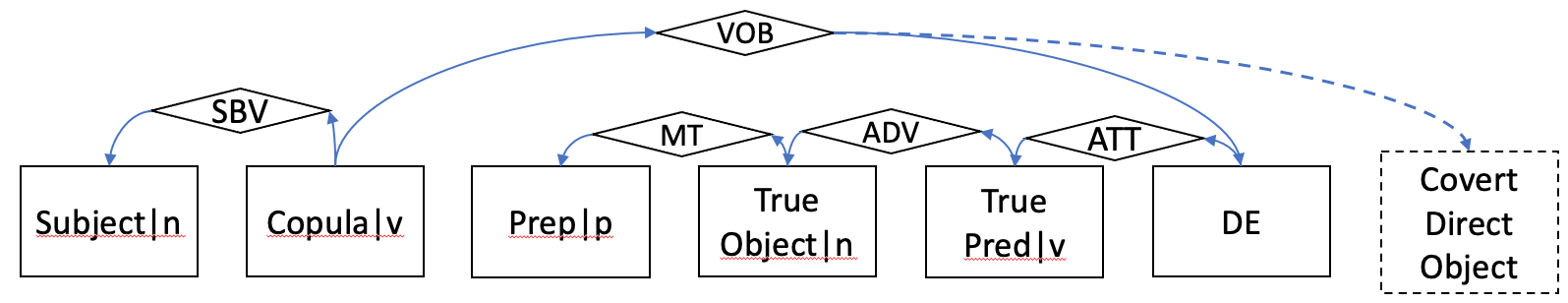}
    \end{minipage}
    \\\cline{2-2}
    & 
    \textbf{Example}: \textit{``玉米(Corn) 是(is) 从(from) 美国(US) 引进(introduce) 的(De); Corn is introduced from US''}
    \\\cline{2-2}
    &
    \textbf{Relation 1}: (Subject, Copula·Prep·X·True\_Pred·DE, True\_Object);\quad (玉米(Corn), 是·从·X·引进·的(is·from·X·introduced·DE), 美国(US))
    \\\hline
    
    \multirow{4}{*}{\textbf{E.2}} &
    \begin{minipage}{.3\textwidth}
      \includegraphics[scale=0.22]{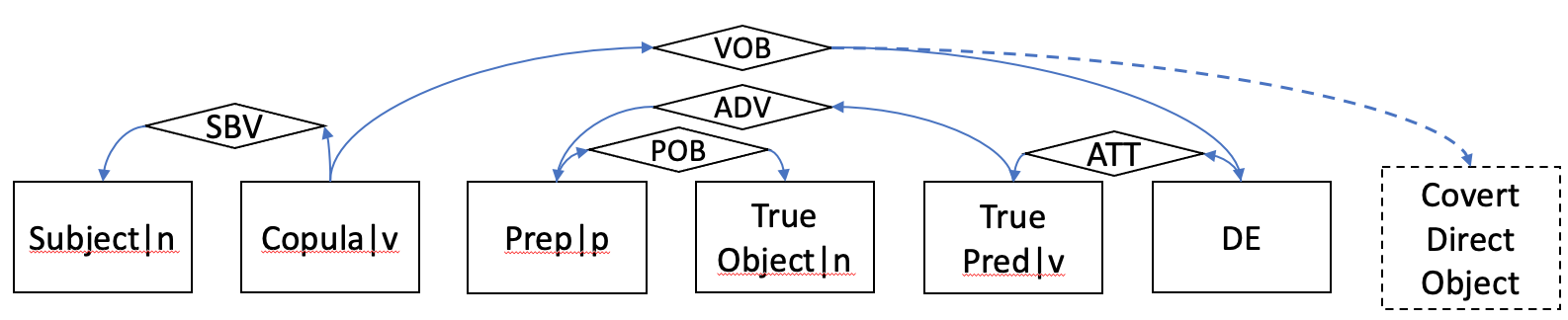}
    \end{minipage}
    \\\cline{2-2}
    & 
    \textbf{Example}: \textit{``设备(device) 是(is) 用(from) 木头(wood) 做(make) 的(De); The device is made of wood''}
    \\\cline{2-2}
    &
    \textbf{Relation 1}: (Subject, Copula·Prep·X·True\_Pred·DE, True\_Object);\quad (设备(device), 是·用·X·做·的(is·from·X·made), 木头(wood))
    \\\hline
    
    \multirow{4}{*}{\textbf{E.3}} &
    \begin{minipage}{.3\textwidth}
      \includegraphics[scale=0.22]{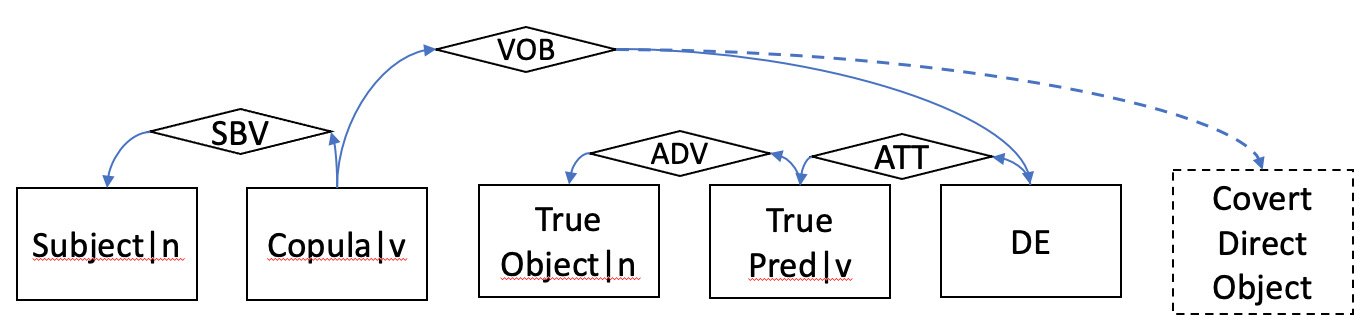}
    \end{minipage}
    \\\cline{2-2}
    & 
    \textbf{Example}: \textit{``设备(device) 是(is) 木头(wood) 做(make) 的(De); The device is made of wood''}
    \\\cline{2-2}
    &
    \textbf{Relation 1}: (Subject, Copula·X·True\_Pred·DE, True\_Object);\quad (设备(device), 是·X·做·的(is·X·made), 木头(wood))
    \\\hline
    
    \multirow{4}{*}{\textbf{E.4}} &
    \begin{minipage}{.3\textwidth}
      \includegraphics[scale=0.22]{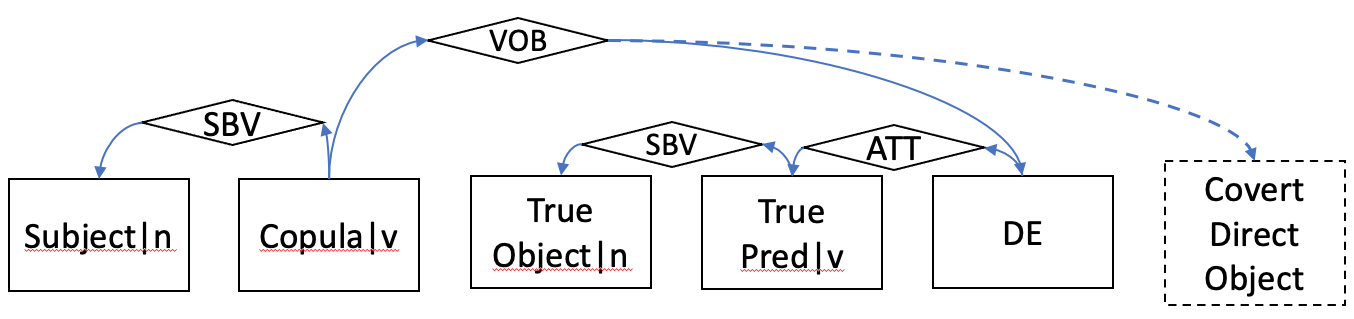}
    \end{minipage}
    \\\cline{2-2}
    & 
    \textbf{Example}: \textit{``设备(device) 是(is) 木匠(carpenter) 做(make) 的(De); The device is made by a carpenter''}
    \\\cline{2-2}
    &
    \textbf{Relation 1}: (Subject, Copula·X·True\_Pred·DE, True\_Object);\quad (设备(device), 是·X·做·的(is·X·made·DE), 木匠(carpenter))
    \\\hline
    
  \end{tabular}
  \caption{Yet more syntactic analysis in \S\ref{Sec:P4CORE} illustrated with diagrams, examples and their extracted relations.}\label{Tab:OREDiagrams3}
\end{table*}

\end{CJK*}
\end{document}